# Large Language Models Meet Biomedical Knowledge Graphs for Mechanistically Grounded Therapeutic Prioritization

Chih-Hsuan Wei[1*], Chi-Ping Day[2*], Zhizheng Wang[1*], Christine C. Alewine[3], Betty Tyler[4], Hasan Slika[4], David Saraf[4], Chin-Hsien Tai[5], Joey Chan[1], Robert Leaman[1] and Zhiyong Lu[1†]

*[1]Division of Intramural Research (DIR), National Library of Medicine (NLM), National Institutes of Health (NIH); Bethesda, MD 20894, USA*

*[2]Cancer Data Science Laboratory, Center for Cancer Research, National Cancer Institute (NCI), National Institutes of Health (NIH), Bethesda, MD 20894, USA*

*[3]Dartmouth Health, Dartmouth Hitchcock Medical Center, Lebanon, NH, USA*

*[4]Hunterian Neurosurgical Laboratory, Johns Hopkins School of Medicine, Baltimore, MD 21231, USA*

*[5]National Cancer Institute (NCI), National Institutes of Health (NIH), Bethesda, MD 20894, USA*

[*]Equal contribution. [†]To whom correspondence should be addressed: zhiyong.lu@nih.gov

## Abstract

Drug repurposing is often framed as a candidate identification task, but existing approaches provide limited guidance for distinguishing biologically plausible candidates from historically well-connected ones. Here we introduce DrugKLM, a hybrid framework that integrates biomedical knowledge graph structure with large language model-based mechanistic reasoning to enable mechanistically grounded therapeutic prioritization. Across benchmark datasets, DrugKLM outperforms knowledge graph–only and language model-only baselines, including TxGNN. Beyond improved recall, DrugKLM confidence scores exhibit functional alignment with molecular phenotypes: higher scores are associated with transcriptional signatures linked to improved survival across 12 TCGA cancers. The scoring framework preferentially captures biologically perturbational signals rather than historical indication patterns. Expert curation across five cancers further reveals systematic differences in prioritization behavior, with DrugKLM elevating candidates supported by coherent mechanistic rationale and disease-specific clinical context. Together, these results establish DrugKLM as an evidence-integrative framework that translates heterogeneous biomedical data into mechanistically interpretable and clinically grounded therapeutic hypotheses.

## Introduction

Drug repurposing - the identification of new therapeutic uses for existing drugs - offers a practical strategy to accelerate drug development. However, identifying clinically meaningful repurposing opportunities remains challenging because biomedical knowledge is vast, heterogeneous, and distributed across structured databases (e.g., drug–gene associations, ontologies), unstructured literature (e.g., PubMed abstracts, clinical trial reports), high-dimensional molecular datasets, and diverse clinical outcomes.

In precision medicine, computational repurposing strategies typically rely on one of several paradigms: knowledge graph (KGs) inference[1-3], molecular similarity[4,5], perturbational signatures[6,7], large language models (LLMs)[8,9], or hybrid approaches[10-16]. KG-based systems exploit structured biological networks to infer candidates from topological relationships among drugs, genes, and diseases. For example, the recent TxGNN[3] system leverages graph neural networks over large-scale biomedical knowledge graphs to prioritize drug–disease associations through inductive link prediction based on relational connectivity patterns. As such, TxGNN sets a new state of the art in drug repurposing prediction. In contrast, LLM-based pipelines synthesize unstructured textual knowledge to generate therapeutic hypotheses through free-text reasoning.

Despite demonstrated utility, most existing methods emphasize a limited subset of available evidence. KG-based approaches may underutilize rapidly evolving textual knowledge dispersed the literature, whereas LLM-based approaches may lack explicit grounding in the relational structure of large-scale biological networks. This separation can limit interpretability, mechanistic transparency, and clinical alignment. To address this gap, we introduce DrugKLM, a hybrid framework that unifies structured knowledge with mechanistic LLM reasoning to both generate candidate therapies and evaluate them through an interpretable scoring function. DrugKLM integrates evidence from KG subgraphs, literature-derived associations, gene–drug–disease relationships, and pathway-level perturbational signatures. These signals are synthesized through a structured chain-of-thought (CoT) process that yields an interpretable confidence score for each drug-disease pair prediction.

We evaluate whether our scoring function captures clinically meaningful signals using complementary quantitative and expert-driven analyses. First, we assess survival relevance by testing whether higher DrugKLM scores correspond to transcriptional signatures associated with improved patient survival across multiple The Cancer Genome Atlas (TCGA) cancer cohorts[17]. Second, we conduct expert curation of top-ranked predictions across five cancers, examining alignment with disease-specific clinical evidence from ClinicalTrials.gov[18] and the biomedical literature, and comparing ranking behavior with TxGNN. To demonstrate its real-world utility, we further present a melanoma case study where DrugKLM supports subtype candidate prioritization and mechanistic hypothesis generation for downstream experimental and translational investigation.

Together, these analyses demonstrate that DrugKLM integrates heterogeneous biomedical evidence into a scoring framework that not only combines knowledge graph structure with mechanistic LLM reasoning, but also enables explicit tracing of supporting evidence across drug-, gene-, and pathway-level signals. By systematically evaluating the relative strength and limitations of distinct evidence types within a unified scoring scheme, DrugKLM provides an interpretable prioritization strategy aligned with both mechanistic reasoning and clinical signals. The breadth of quantitative, survival-based, clinical-trial, and expert-curated evaluations further establishes a comprehensive validation paradigm that may serve as a reference framework for future computational drug repurposing studies.

## Results

### Overview of DrugKLM and Its Clinically Validated Multi-Evidence Scoring Framework

DrugKLM first standardizes the input disease description to support downstream knowledge graph (KG) inference and evidence integration (Fig. 1a). The raw disease description text is parsed to extract the disease name and genomic variant. The disease name is linked to a disease entity in the KG by exact string match; if no exact match is available, the closest parent or related entity is selected based on semantic similarity computed by MedCPT, a state-of-the-art biomedical text encoder. The original disease description is retained to condition subsequent language model–based candidate generation and scoring, while the mapped KG entity is used for graph-based retrieval.

DrugKLM then constructs a high-coverage candidate pool (Fig. 1b) by integrating structured graph inference with mechanistic language model reasoning. Drugs are ranked using two Hierarchy-Aware Knowledge Graph (HAKE)[19] embedding models trained on PrimeKG[20]: a baseline structural model (KGE) and a literature-weighted variant (KGwE) that incorporates publication-derived edge evidence. KGE relies solely on graph topology, whereas KGwE weights relationships by supporting literature. Using both models balances structurally inferred associations with evidence-supported relationships.

In parallel, DrugKLM applies a two-step prompting strategy. First, the model decomposes the disease into mechanistic drivers, including molecular and pathophysiological processes (e.g., dysregulated pathways, causal genes, disrupted biological processes). Second, it proposes drugs that target those mechanisms. Together, these channels provide complementary strengths: KG-based models capture latent topological structure in biomedical knowledge graphs, while the LLM contributes context-dependent mechanistic reasoning.

For each disease–drug pair (Fig. 1c), DrugKLM assembles a multi-evidence profile from three perspectives:

- Drug evidence: KG subgraphs, FDA label statements, and literature snippets.
- Gene evidence: literature-based annotations[21,22] that capture gene–disease and gene–drug associations, along with gene-level support derived from KGE.
- Pathway evidence: drug perturbational expression signatures from SigCom-LINCS [23] integrated with IC50 information from GDSC (Genomics of Drug Sensitivity in Cancer)[24] to construct up- and down- regulated gene sets aligned with disease pathways.

This multi-evidence profile is processed through a structured CoT framework that highlights supporting facts (Fig. 1d), explicitly weighs risks and limitations, and assigns a 0–100 confidence score using a fixed scoring guideline. These criteria were developed in consultation with domain experts to identify key factors relevant to drug–disease evaluation and are explicitly encoded in the evaluation prompt, to ensure adherence to a predefined, expert-informed process.

In the final stage (Fig. 1e), a GPT-based model integrates FDA approval status, ClinicalTrials.gov records, and disease-specific literature to assign each drug to a standardized clinical stage. To focus on candidates of practical interest, DrugKLM applies two filters: (1) restricting to FDA-approved drugs to enable repurposing, and (2) prioritizing clinical stages most relevant to researchers and clinicians.

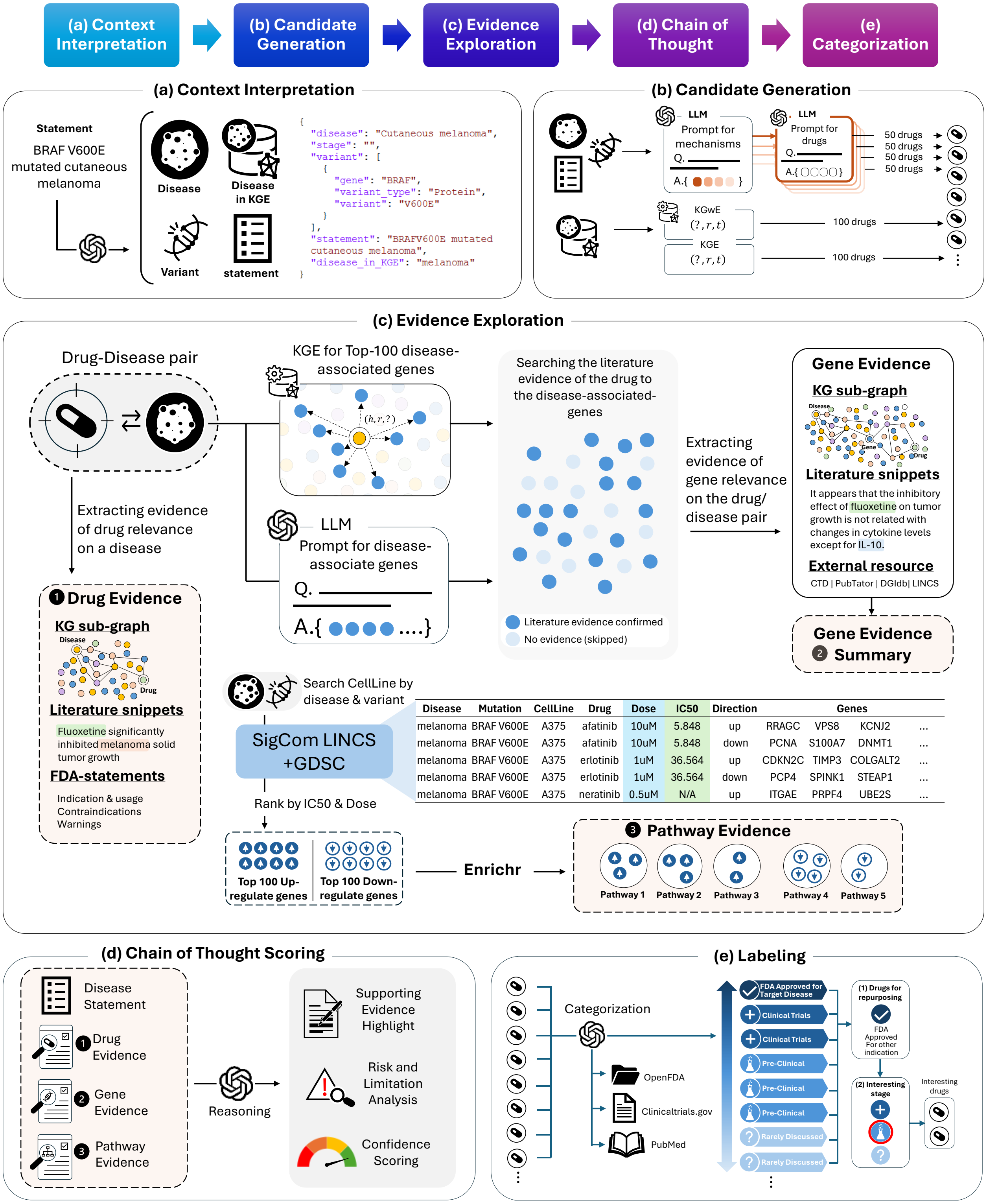

Fig. 1 The DrugKLM framework for drug repurposing and clinically validated multi-evidence scoring. (a) Context interpretation: The LLM standardizes the disease query by extracting the full disease name, subtype, and relevant variants, and maps the refined description to the most appropriate KG entity using

MedCPT. (b) Candidate generation: HAKE-based KG embeddings combined with a mechanism-aware LLM propose mechanism-aligned candidates. (c) Evidence exploration: Drug, gene, and pathway evidence are integrated from KG subgraphs, curated resources, FDA statements, and LINCS expression signatures. (d) Chain-of-thought scoring: A fixed scoring criteria synthesizes evidence and assigns confidence scores. (e) Categorization: GPT-based integration of regulatory status, clinical trial records, and literature evidence assigns candidate drugs to standardized clinical stages.

**Recall Evaluation on PharmacotherapyDB**

Recall evaluation assesses how effectively DrugKLM retrieves known therapeutic relationships from benchmark datasets, providing a direct measure of its ability to recover clinically validated drug-disease associations. We evaluated recall using PharmacotherapyDB (https://github.com/dhimmel/indications), a physician-curated catalog of drug therapies, covering 91 diseases and 1,145 therapeutic indications (excluding 243 non-indications), widely used as a benchmark in drug-indication studies[25,26].

To isolate examine the contributions of language-based and knowledge graph–based reasoning, we compared six model configurations on PharmacotherapyDB (Fig. 2a). A single-prompt GPT-4.0 model (LLM) achieved the lowest recall (14.15%). In contrast, our two-stage prompting strategy (LLM+) substantially improved recall to 60.26%, indicating that mechanism-level decomposition can enhance LLM inference.

We next evaluated three knowledge graph embedding configurations. The base HAKE model (KGE), trained on PrimeKG[20], recovered 57.03% of known disease-drug links by leveraging hierarchical structure in the KG. The evidence-weighted variant (KGwE), which incorporates literature-derived edge weights from PubTator[22], achieved higher recall (62.18%) by emphasizing experimentally or clinically supported relationships. Combining both embeddings (KGEUKGwE) further increased recall to 67.86%, outperforming either model alone.

Fig. 2b shows the distribution of therapeutic indications recovered by LLM+, KGE, and KGwE. KGE and KGwE models produce closely related but complementary results, indicating that literature weighting provides additional gains beyond structural embeddings alone. In contrast, the overlap between the KG-based models and LLM+ is smaller, with many indications uniquely identified by one approach. These results demonstrate that KG embeddings and LLM-based reasoning capture largely distinct therapeutic signals and are therefore complementary within DrugKLM.

Because pretrained weights for TxGNN are not publicly available, direct benchmarking was not feasible. Instead, we queried the TxGNN web interface for 13 diseases (10 cancers and 3 chronic diseases), included in PharmacotherapyDB. As shown in Fig. 2c, DrugKLM achieved 78.47% recall compared with 61.24% for TxGNN (diseases listed in Supplementary Table S1).

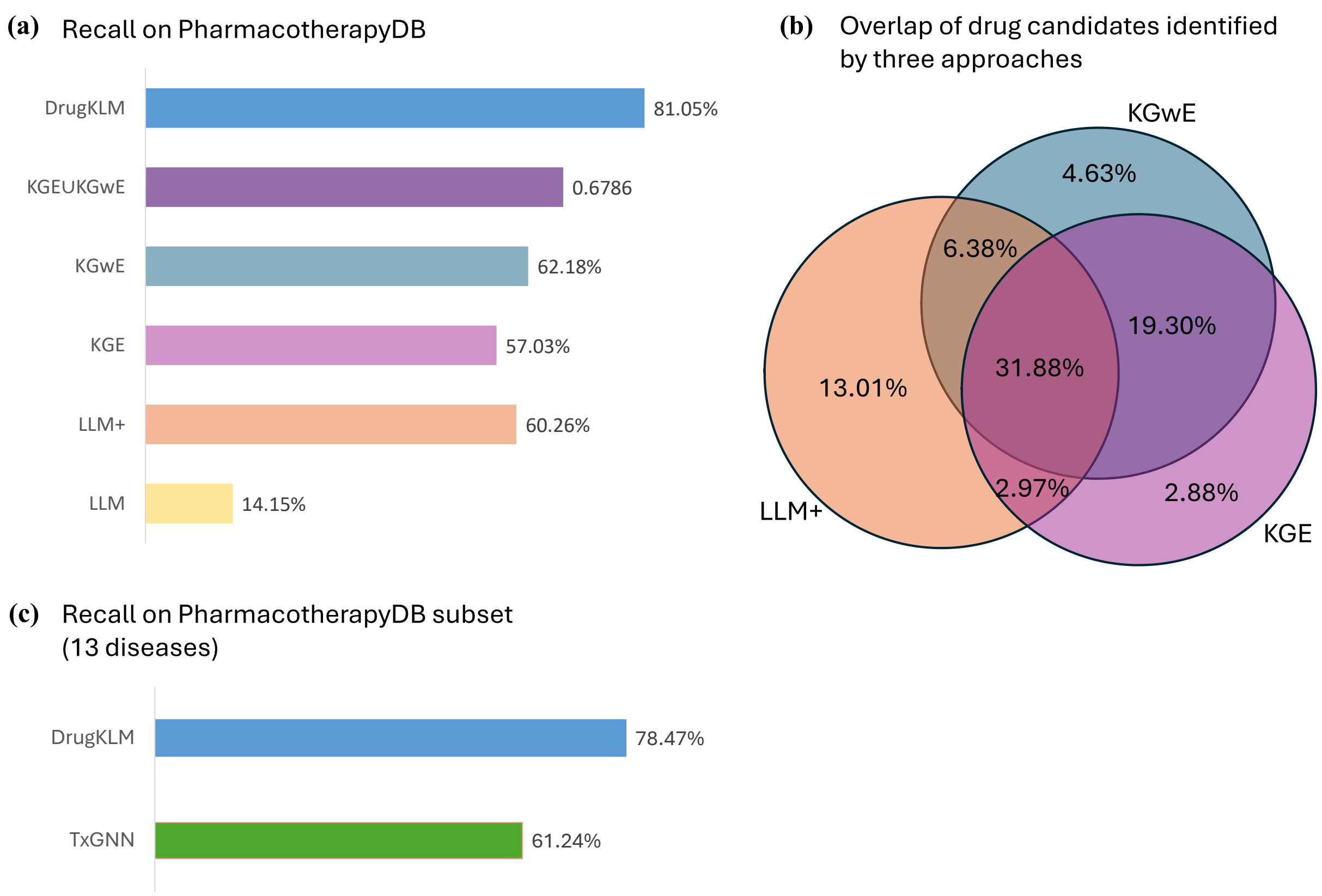

Fig. 2 Recall evaluation and clinical-trial alignment across benchmark and ClinicalTrials.gov analysis. (a) Recall on PharmacotherapyDB across six configurations, illustrating the contributions of mechanistic prompting (LLM+), knowledge graph embeddings (KGE, KGwE, KGEUKGwE), and the full DrugKLM. DrugKLM achieves the highest recall. (b) Venn diagram showing overlap among LLM+, KGE, and KGwE. Percentages on PharmacotherapyDB indicate the proportion of indications uniquely or jointly recognized by each method. Percentages sum to 81.05%, corresponding to the overall recall achieved by DrugKLM in panel (a). (c) Recall comparison between DrugKLM and TxGNN on 13 diseases available through the TxGNN web interface. DrugKLM demonstrates substantially higher recall.

**Clinical Validation of DrugKLM Scores: Survival-Relevance**

To evaluate whether DrugKLM scores reflect mechanistic and clinical relevance, we computed confidence scores for 150 drugs per a given indicated cancer and examined whether higher scores were associated with clinical efficacy. Drug perturbational gene expression (DPGE) signatures from SigCom-LINCS were used as surrogate markers of drug-induced effects in cancer cells. We then assessed whether expression of these DPGE signatures in tumors was associated with patient survival across multiple TCGA cancer cohorts.

We collected bulk transcriptomic data of 12 cancer types (Supplementary Table S1) from TCGA. For each drug-disease pair, enrichment of the corresponding DPEG signature was computed for each tumor within the indicated cancer type. Patients were stratified by enrichment scores, and survival analysis was performed to estimate hazard ratios (HR) (see Methods). An HR > 1 indicates that the drug-induced signature is associated with reduced survival probability, and vice versa. We computed the Spearman correlation between DrugKLM confidence scores and HRs to assess alignment. Because HR estimates for sparsely represented drug–disease pairs are unstable, analysis was restricted to pairs with adequate sample representation and observed survival events.

Across the 12 cancers, the pooled Spearman correlation between HR and DrugKLM score was negative and statistically significant ($r = -0.149$, $p = 0.0107$; Fig. 3a), indicating that higher DrugKLM scores are associated with putative drug efficacy that improved survival in aggregate analyses. In contrast, TxGNN showed a positive but non-significant correlation ($r = 0.037$, $p = 0.7863$; Supplementary Fig. S1a). As expected, correlation magnitude and direction varied across individual cancers, particularly those with fewer evaluable pairs.

We next compared DrugKLM and TxGNN directly across survival evidence (Fig. 3b). Each drug–disease pair is plotted with its TxGNN score (x-axis) and DrugKLM score (y-axis), colored by TCGA-derived hazard ratio. Pairs associated with better survival tend to cluster at higher DrugKLM scores, indicating closer alignment with survival data relative to TxGNN.

To quantify the contribution of individual evidence sources, we performed leave-one-evidence-out ablation analyses, removing each evidence component and re-running the full scoring pipeline. We also evaluated a scoring-paradigm ablation in which the structured chain-of-thought module was replaced with a fixed, rule-based scoring scheme while keeping the same evidence inputs. As shown in Fig. 3c, removal of any single evidence source consistently reduced alignment with TCGA survival outcomes indicated by HR, demonstrating that DrugKLM benefits from complementary evidence integration.

Pathway evidence ablation produced the largest reduction in survival correlation, whereas its impact on ClinicalTrials.gov alignment was comparatively modest (Fig 3c and S2c). The scoring prompt for automated ClinicalTrials.gov–based relevance evaluation is shown in Fig. S4. This pattern is expected, as gene set enrichment analysis (GSEA) captures pathway-level biological signals that are more directly reflected in pharmacological action and therefore survival associations, as compared to the downstream clinical-trial outcomes. In contrast, gene- and drug-level evidence contributed more strongly to clinical-trial alignment, consistent with their proximity to therapeutic indication. Together, these findings indicate that DrugKLM integrates evidence across multiple pharmacological and clinical resolutions.

Replacing CoT-based scoring with a structured, criteria-based scheme led to a substantially larger reduction in correlation than removal of any individual evidence source. The non-CoT baseline applied a domain-informed evaluation prompt (Supplementary Fig. S3) with fixed scoring dimensions but suppressed intermediate reasoning steps. The resulting performance gap highlights the importance of reasoning-enabled evidence integration - rather than differences in scoring criteria or domain knowledge - in driving DrugKLM's clinical alignment.

**Human Expert Curation Reveals Systematic Differences in Ranking Behavior between DrugKLM and TxGNN**

To evaluate whether model rankings align with expert clinical reasoning beyond benchmark recall and trial coverage, we conducted human curation of the top 30 drugs ranked by DrugKLM and TxGNN across five cancers: liver cancer, colon cancer, pancreatic cancer, melanoma, and glioblastoma. Curation was performed by clinicians and cancer researchers with formal biomedical training and experience evaluating clinical evidence. All curators followed a unified annotation guideline, specifying standardized criteria for clinical evidence status, FDA approval context, and treatment modality, as detailed on our GitHub repository. Each drug–disease pair was manually reviewed and categorized according to the highest level of clinical evidence achieved for the specified disease, irrespective of whether evidence arose from monotherapy or combination therapy. FDA approval status for other indications and treatment modality were also recorded. Curated raw data is provided in Supplementary Table S2.

As shown in Fig. 3d, DrugKLM and TxGNN exhibit distinct distributions of clinical evidence levels among their top-ranked predictions. DrugKLM more frequently prioritizes drugs supported by positive mechanistic or clinical evidence, reflecting a stronger emphasis on biologically grounded and clinically actionable candidates. In contrast, TxGNN includes a higher proportion of candidates less frequently discussed in disease-specific clinical contexts and supported by fewer focused studies, consistent with its design goal of prioritizing candidates for rare or under-studied diseases.

Analysis of curated drug classes further reveals systematic differences in the therapeutic classes (Fig. 3e). Across the five cancers, DrugKLM predictions are enriched for targeted small-molecule therapies and immunotherapies, whereas TxGNN predictions include a larger fraction of cytotoxic chemotherapies. DrugKLM preferentially ranks pathway-specific, mechanism-aligned agents, such as BRAF inhibitors (e.g., vemurafenib) for V600E-mutated melanoma – whose identification requires explicit disease–gene–pathway reasoning and cannot be recovered reliably through drug–drug similarity or indication co-occurrence alone. In contrast, TxGNN more frequently prioritizes cytotoxic agents with broad historical use across cancer types (e.g., lomustine and carmustine) which exhibit high connectivity in knowledge graphs and literature. Notably, many of these broadly used chemotherapies are also identified by DrugKLM but are assigned lower confidence scores and ranked below mechanism-aligned candidates, reflecting DrugKLM's preference for disease-specific biological relevance over historical prevalence.

We additionally performed an automated ClinicalTrials.gov–based relevance analysis (Supplementary Materials). Because this analysis relies on an AI reviewer, it is not treated as primary clinical validation. However, the resulting ranking patterns were consistent with human curation: higher-ranked DrugKLM candidates were more frequently associated with stronger trial-based relevance than those prioritized by TxGNN (Supplementary Fig. S2).

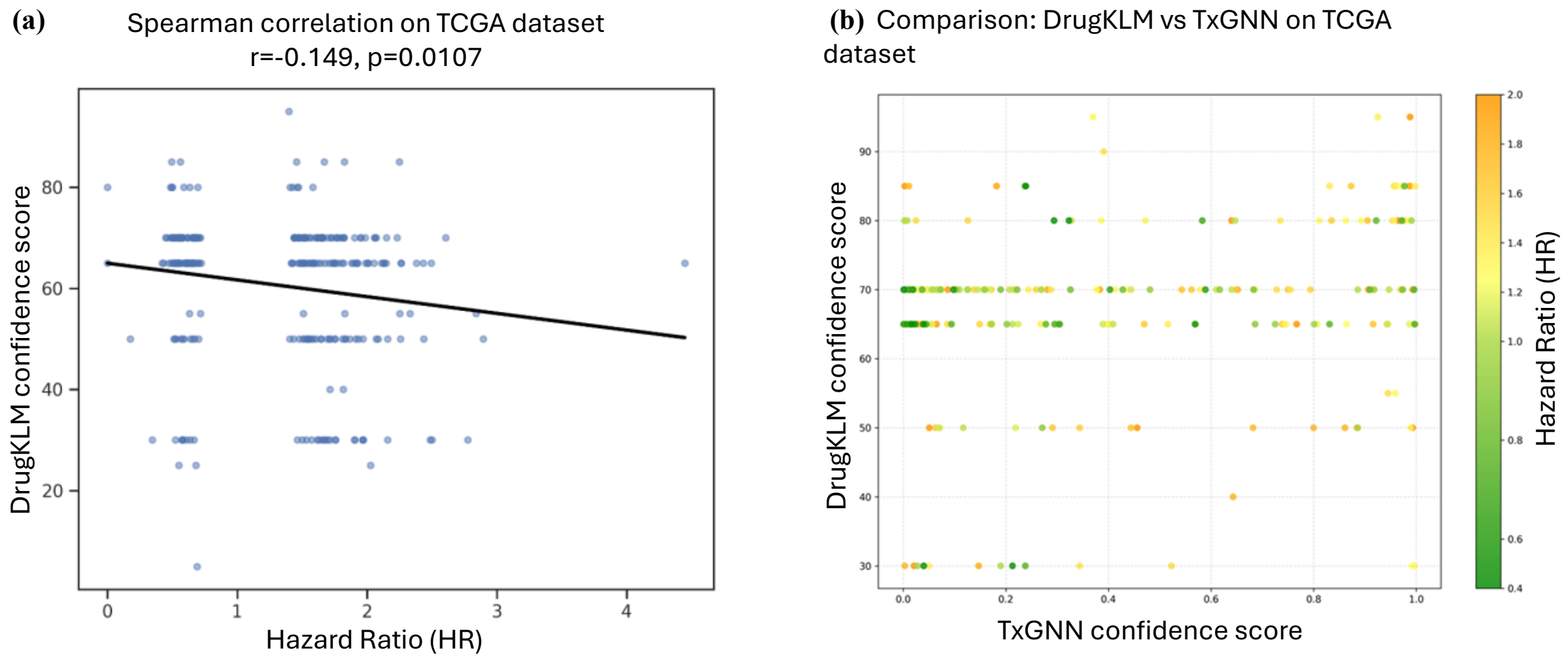

**(c)** Leave-one-evidence-out ablation analysis on TCGA dataset

| | Spearman r | p-value |
|---|---|---|
| DrugKLM | -0.149 | 0.0107 |
| DrugKLM w/o Drug Evidence | -0.118 | 0.0447 |
| DrugKLM w/o Gene Evidence | -0.110 | 0.0496 |
| DrugKLM w/o Pathway Evidence | -0.092 | 0.0106 |
| DrugKLM w/o CoT (using heuristic rule for scoring) | -0.031 | 0.0641 |

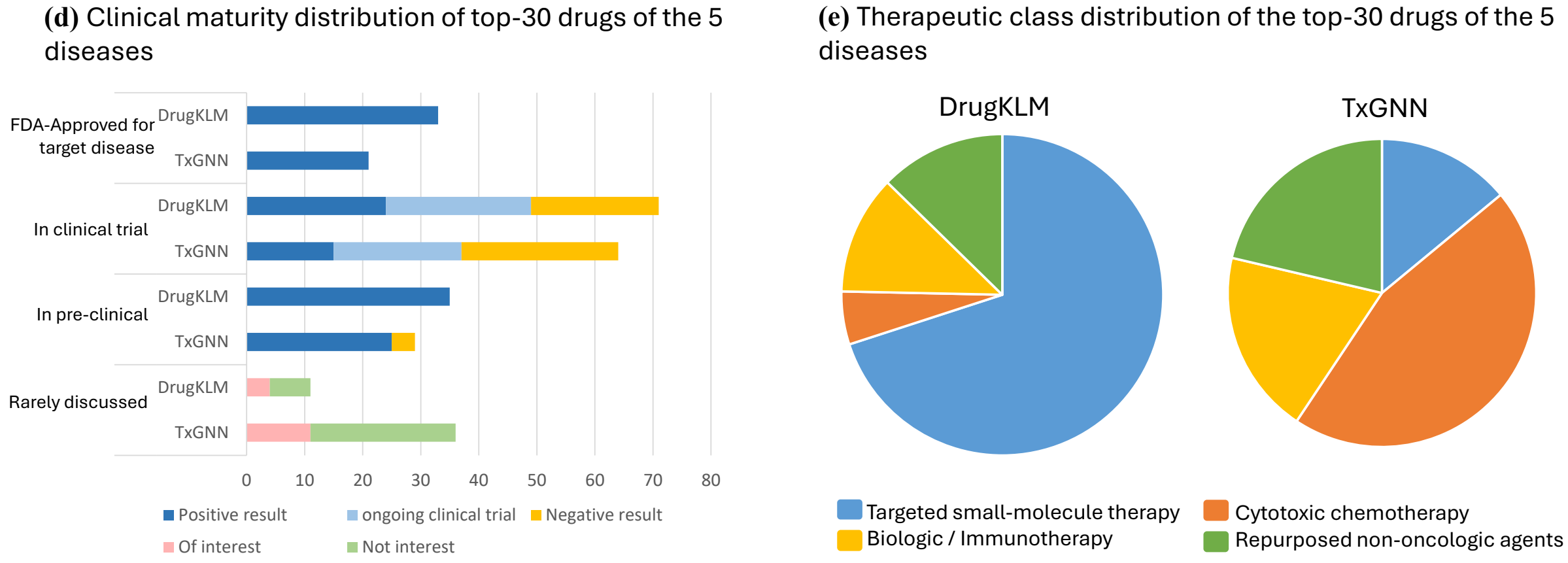

Fig. 3. Clinical validation of DrugKLM confidence scores across survival and expert curation. (a) Relationship between DrugKLM confidence scores and hazard ratios (HRs) derived from TCGA transcriptional signatures across 12 cancers. Each point represents a drug–disease pair with sufficient sample support. Higher DrugKLM scores correspond to lower HRs, indicating alignment with improved survival. (b) Direct comparison of DrugKLM and TxGNN scores for drug–disease pairs with TCGA survival data. Points are colored by HR. More favorable survival outcomes cluster at higher DrugKLM scores. (c) Leave-one-evidence-out and scoring-paradigm ablation analysis. Each evidence component (drug, gene, or pathway) was removed in turn, and the full pipeline was re-run. A separate ablation replaced chain-of-thought (CoT) scoring with a fixed rule–based scheme using identical evidence inputs. Spearman correlations with TCGA survival and ClinicalTrials.gov endpoints are reported. (d) Distribution of clinical evidence levels among the top30 ranked drugs, categorized by positive outcomes, ongoing trials, and negative results. (e) Therapeutic class distribution of the top30 drugs ranked by DrugKLM and TxGNN across five cancers.

**Use Case Study: Atovaquone for Melanoma**

DrugKLM generates ranked confidence scores for hundreds of drug–disease pairs, but practical application requires translating these associations into biologically meaningful hypotheses. We therefore present a melanoma case study illustrating how DrugKLM supports candidate selection by integrating subtype-specific confidence profiling with global confidence patterns.

Melanoma is classified into four major subtypes by occurring sites- cutaneous, acral, mucosal, and uveal. Although all four derived from melanocytes, and share core developmental programs and driver pathways, they exhibit distinct pathological and molecular features. Because cutaneous melanoma accounts for over 90% of all melanoma cases, most therapies were developed for this subtype and subsequently applied to others, despite heterogeneous treatment responses. Deciphering therapies with consistent or subtype-specific efficacy remains an urgent clinical need for melanoma treatment[27].

To address this issue, we applied DrugKLM to all four melanoma subtypes. For drugs detected across subtypes, we constructed a four-dimensional confidence profile and applied principal component analysis (PCA) to capture patterns of cross-subtype consistency and specificity (Fig. 4a). Based on the distribution of scores among the four melanoma subtypes, we identified five clusters associated with distinct indications, targets, and clinical development stages (Supplementary Table S6). Cluster 1 contained drugs with consistently high and balanced scores across all subtypes, suggesting broadly applicable relevance rather than subtype-restricted efficacy.

To prioritize candidates within Cluster 1, we applied the categorization procedure described in Fig. 1e, integrating FDA approval status with melanoma-specific evidence from ClinicalTrials.gov and the literature. This step filtered for drugs approved for non-melanoma indications but supported by positive preclinical evidence in melanoma. Atovaquone emerged as the top-ranked candidate meeting these criteria, with high scores across all four subtypes.

Atovaquone targets mitochondrial complex III and is FDA-approved for parasitic infection. DrugKLM predicts potential repurposing for melanoma, particularly combination therapy. Figure 4b-d summarizes the model-generated mechanistic rationale and supporting evidence. The KG subgraph (Fig. 4b) links atovaquone to melanoma biology through mitochondrial respiration, MAPK signaling pathways, immune-related apoptosis, and drug transport processes. LINCS-based gene set enrichment analysis (Fig. 4c) shows enrichment of immune and metabolic pathways, including interferon-gamma signaling, complement activation, neutrophil activation, and oxidative stress pathways implicated in melanoma response to immune checkpoint blockade (ICB)[28,29]. The integrated prediction summary (Fig. 4d) consolidates DrugKLM score, regulatory status, combination potential, mechanistic rationale, and suggested experimental directions. Together, these outputs explain why atovaquone was prioritized based on consistent cross-subtype score and mechanistic support.

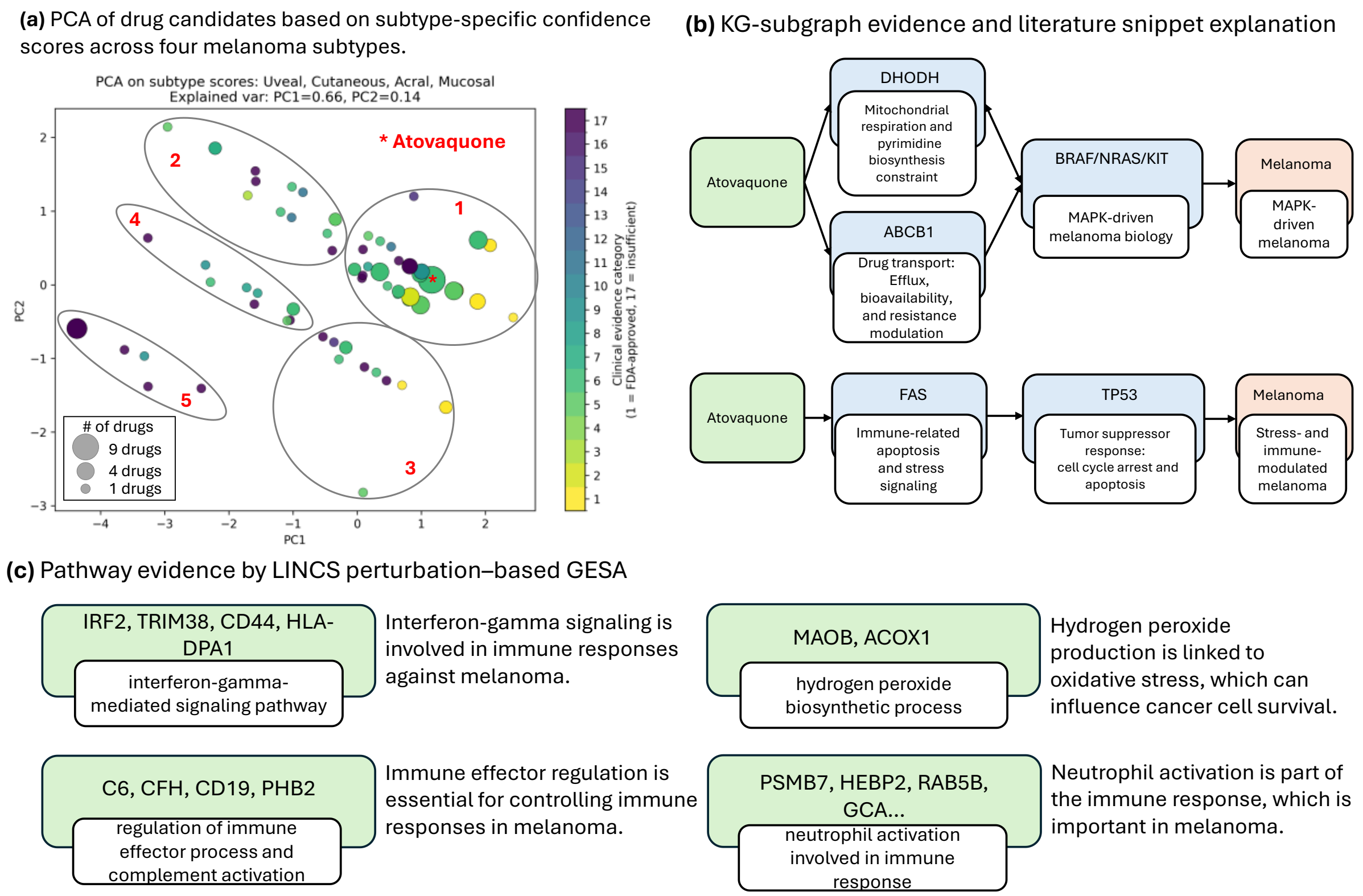

**(d)** Output of prediction

| Attributes | Results |
|---|---|
| Confidence level (score) | Moderate High (70) |
| FDA status | FDA-approved for non-melanoma indications: malaria; Pneumocystis pneumonia prophylaxis |
| Combination or mono therapy | Combination therapy |
| Reference | PMID:39652639 |
| Rationale bullets | Atovaquone inhibits mitochondrial electron transport chain complex III, disrupting oxidative phosphorylation and reducing ATP production, which can impair melanoma cell metabolism. GSEA pathways such as “regulation of complement activation” and “regulation of immune effector process” are upregulated, suggesting immune modulation relevant to melanoma. |
| Next steps | Conduct in vitro assays to evaluate the effect of Atovaquone on melanoma cell lines with BRAF and NRAS mutations.<br>Investigate the impact of Atovaquone on immune response pathways in melanoma, particularly those identified in GSEA (e.g., regulation of complement activation, interferon-gamma-mediated signaling). |

Fig. 4 Integrated prioritization and mechanistic characterization of a melanoma drug candidate using DrugKLM. (a) Principal component analysis (PCA) of subtype-specific DrugKLM confidence scores across four melanoma subtypes (cutaneous, uveal, acral, and mucosal), providing a global overview of cross-subtype confidence patterns among candidate drugs. Each circle represents one or more drugs; size indicates count, color denotes dominant clinical evidence category. (b) Knowledge graph subgraph linking atovaquone to melanoma-related genes and pathways. (c) LINCS–based pathway enrichment of atovaquone-induced transcriptional changes. (d) Integrated prediction summary combining multi-level evidence.

## In Silico Validation of Atovaquone Efficacy

Guided by DrugKLM's reasoning, we performed virtual efficacy analyses of atovaquone and in combination with anti-PD-1 therapy. We defined a target-gene signature comprising mitochondrial complex III subunits (CYC1, UQCRFS1, and MT-CYB[30]). Following prior studies, target-gene expression was used as a surrogate for functional activity[31]; lower expression simulates pharmacologic inhibition.

In TCGA melanoma data, patients stratified by signature expression showed improved survival with lower signature levels (HR = 0.56, Fig. 5a), consistent with potential therapeutic effect.

To evaluate combination therapy, we analyzed GSE91061[32], which includes pre- and on-treatment transcriptomes from melanoma patients receiving nivolumab (anti-PD-1). Lower signature expression was associated with improved survival in pre-treatment samples (HR = 0.47, p = 0.0746, Fig. 5b), and showed a stronger, statistically significant association in on-treatment samples (HR = 0.41, p = 0.0475; Fig. 5c), suggesting atovaquone may enhance ICB therapy.

We further examined context dependence using GSE78220[33], an independent pre-treatment melanoma cohort. Immune activation was inferred from combined GZMA and PRF1 expression[34], samples were categorized into immune-active, immune-intermediate, and immune-inactive groups. Survival associations were more pronounced in immune-active tumors (HR = 0.45; Fig. 5d) than immune-inactive tumors (HR = 1.26; Fig. 5e), supporting a potential role for atovaquone in augmenting anti-tumor immunity. This finding is consistent with the combination-therapy pattern observed in GSE91061.

Because GSE91061 includes RECIST response data, we performed receiver operating characteristic (ROC) analysis. In pre-treatment samples, the atovaquone signature yielded an area under the curve (AUC) of 0.687, indicating that lower expression predicted tumor response (complete or partial). In on-treatment samples, the AUC increased to 0.748 (Fig. 5f), further suggesting enhanced response potential in combination with anti-PD-1 therapy. Finally, we examined atovaquone across nine cancer types to evaluate the relationship between DrugKLM confidence scores and disease-specific hazard ratios. Higher DrugKLM scores were consistently associated with lower hazard ratios across cancers, (Figure 5g), indicating concordance between predicted confidence and survival benefit. Taken together, these analyses support atovaquone as a mechanistically grounded, cross-subtype melanoma candidate and illustrate how DrugKLM integrates multi-level evidence to generate clinically interpretably hypotheses.

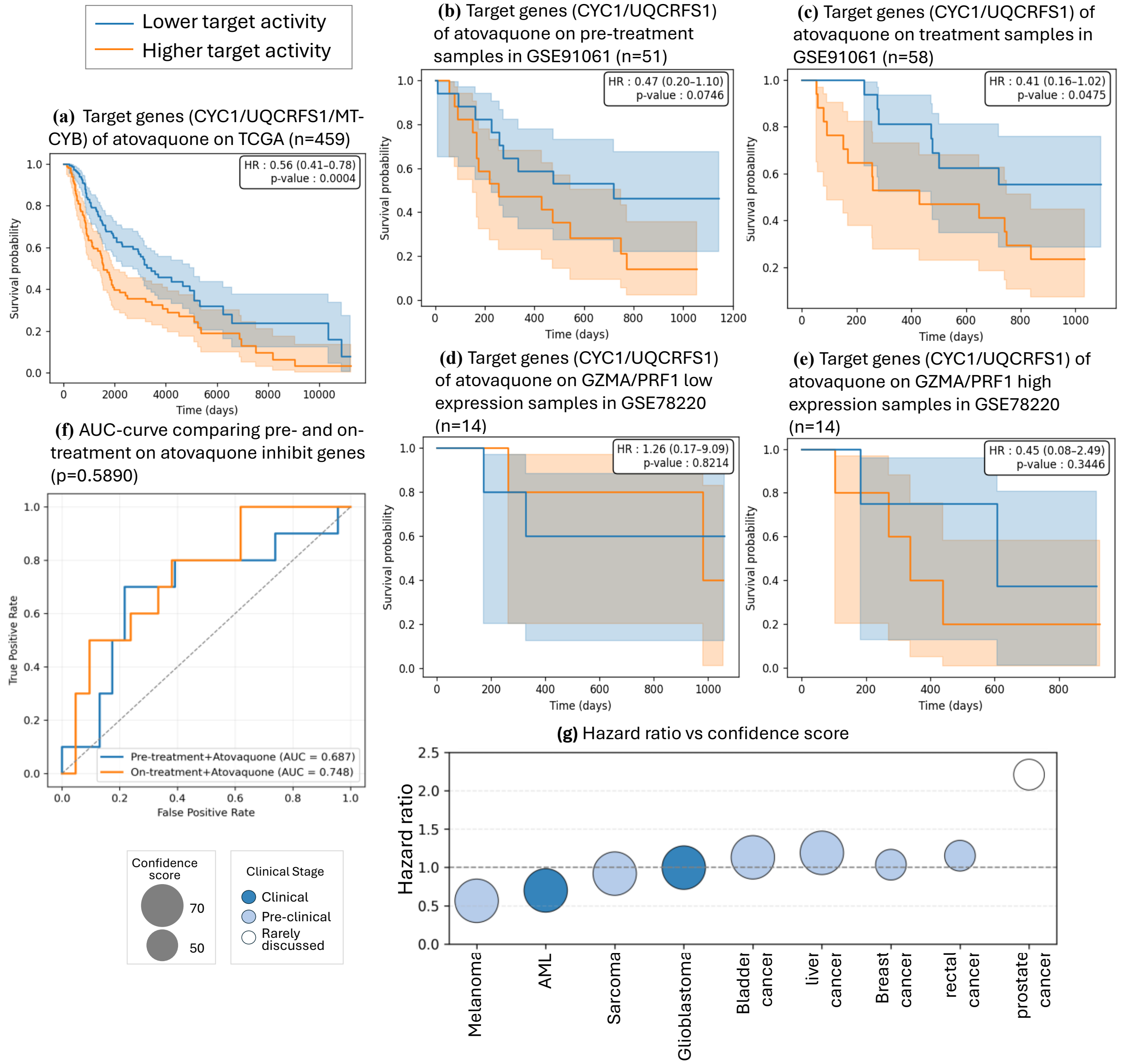

Fig. 5 In silico validation of atovaquone in melanoma. (a) TCGA Kaplan–Meier survival analysis using the atovaquone target –gene signature. (b,c) Kaplan–Meier analyses in GSE91061 (pre- and on-treatment). (d,e) Kaplan–Meier analyses in GSE78220 stratified immune activation status. (f) ROC–AUC comparison prediction in in pre- and on-treatment settings. (g) Association between DrugKLM scores and hazard ratios across cancer types.

## Discussion

In this work, we address a central challenge in computational drug repurposing: whether model-derived confidence scores reflect biologically and clinically meaningful prioritization. We introduce DrugKLM as a unified framework that integrates structured biomedical knowledge with mechanism-aware large language model (LLM) reasoning to evaluate such scores. Across benchmark recall evaluations, DrugKLM outperforms its individual KG-based and LLM-based components as well as the external TxGNN baseline, demonstrating the complementary value of combining relational structure with contextual biological reasoning.

Human curation further indicates that DrugKLM preferentially prioritizes biologically grounded and clinically actionable candidates, particularly in diseases with relatively rich mechanistic and clinical evidence. This behavior reflects DrugKLM's hybrid design: knowledge graph embeddings capture structured relational signals, while LLM-based reasoning contributes disease-specific mechanistic interpretation. The subsequent evidence-aware scoring step then favors candidates supported by coherent biological and clinical evidence. Independent validation strengthens this interpretation: higher DrugKLM confidence scores are associated with survival-linked transcriptional signatures across TCGA cohorts. The melanoma case study further illustrates that DrugKLM can prioritize candidates with potential clinical efficacy before formal evaluation in the specific indication (Fig. 4). Together, these findings suggest that DrugKLM produces rankings grounded in clinical and molecular evidence while remaining interpretable through expert assessment, revealing systematic differences from TxGNN's prioritization patterns.

TxGNN was selected as the external baseline because it is explicitly designed for large-scale drug–disease candidate generation using biomedical knowledge graphs, enabling direct comparison through recall benchmarks and clinical-trial alignment. Other recent LLM- or KG-based therapeutic reasoning systems were not included due to differences in scope, objectives, or scalability. For example, TxAgent[16] emphasizes interactive explanation of predefined drug–disease pairs rather than large-scale candidate generation. K-Path[35], CHANCE[36], and DrugCORpath[37] are designed for task-specific or fixed-vocabulary settings and rely on exhaustive path enumeration or computationally-intensive graph modeling, limiting their applicability to exploratory, large-scale, repurposing analyses.

Agent-based systems such as TheraMind[38] and ESCARGOT[1] are typically evaluated within specific disease domains (e.g., lung cancer or Alzheimer's disease) and incorporate domain-specific assumptions and resources that limit general-purpose, cross-disease evaluation. Given these differences, TxGNN serves as the most appropriate comparator for DrugKLM. Direct quantitative comparison with other systems would therefore not be methodologically appropriate or practically feasible. Supplementary Table S3 summarizes the task formulations of these methods, highlighting differences in user inputs and model outputs that make TxGNN the only directly comparable baseline for DrugKLM.

Supplementary Table S4 further illustrates that most existing approaches rely on a limited subset of evidence modalities. In contrast, DrugKLM integrates multiple complementary evidence sources within a unified scoring framework, enabling more comprehensive and clinically grounded prioritization. Evaluation paradigms in Supplementary Table S5 differ across methods, including zero-shot generalization, EMR-based validation, and wet-experimental validation. These reflect distinct objectives and are not directly comparable to DrugKLM's clinically aligned scoring framework.

Despite these strengths, several limitations should be noted. First, LLM-based clinical trial scoring is not manually adjudicated. Although the scoring prompt was reviewed by domain experts, individual trial assessments are generated automatically, which may limit the capture of nuanced clinical judgment.

Second, the underlying knowledge graph is incomplete. When a queried disease is not explicitly represented in the graph (e.g., “cutaneous melanoma”), the embedding models fall back to the closest parent or related entity (ex., “melanoma”) for candidate retrieval. This approximation can restrict the initial candidate pool. However, DrugKLM does not rely solely on this mapping: the LLM-based candidate generation and scoring stages remain conditioned on the original disease description. Consequently, subtype information and molecular features may still influence prioritization even when the graph representation is coarse.

Third, pathway evidence relies on perturbational signatures from SigCom-LINCS, which are largely derived from cancer cell lines. This limits the informativeness of the pathway module for non-oncology diseases. Nonetheless clinical-trial validation across multiple non-cancer conditions demonstrates consistent performance, suggesting that drug- and gene-level evidence modules remain informative when LINCS-derived pathway data are sparse or unavailable.

Several directions could further strengthen DrugKLM. Expanding the knowledge graph to incorporate more comprehensive disease ontologies, molecular interaction networks, and real-world clinical data would improve candidate coverage and reduce reliance on approximate disease mappings. Integrating additional perturbational datasets beyond cancer-derived LINCS profiles—such as curated supplementary materials or public repositories like GEO—could enhance pathway-level evidence for non-oncology indications. Prospective experimental validation of high-confidence predictions in cellular or animal models will be essential to establish biological relevance. Finally, extending DrugKLM to support combination-therapy reasoning, temporal modeling of disease progression, and individualized patient contexts represents promising directions for increasing translational impact.

## Method

### Context Interpretation

DrugKLM first processes the free-text disease description to extract key identifiers for downstream inference. A dedicated prompt guides the LLM to extract the primary disease name and any mentioned molecular variants (e.g., BRAFV600E) from the user-provided text.

To align the disease with the knowledge graph, all disease entities in PrimeKG are embedded using MedCPT[39] and the extracted disease name is similarly converted into an embedding. If no exact string match exists, the system retrieves the closest disease entity based on embedding similarity. This standardized disease representation is used for knowledge graph-based candidate generation and evidence integration.

For example, if the input specifies provides a granular subtype such as “cutaneous melanoma,” which is not explicitly represented in the knowledge graph, the system maps it to the nearest parent entity “melanoma” in the embedding space. Importantly, the original user-provided disease description is preserved and used to condition downstream LLM-based candidate generation and scoring. This ensures that subtype-specific and molecular context are retained even when the graph mapping is coarse.

## Candidate Generation

*Mechanism-aware LLM*

To complement knowledge graph–based candidate discovery, DrugKLM employs a mechanism-aware large language model (LLM) to propose candidate drugs by decomposing the disease description into mechanistic features, including causal pathways, dysregulated genes, and relevant drug–target interactions. This component captures mechanistic context that may not be fully represented in the knowledge graph and identifies candidates grounded in textual biomedical knowledge.

The mechanism-aware module operates using a two-step prompting strategy. The LLM is first prompted to decompose the disease into its underlying mechanistic features. This produces a structured mechanism profile summarizing key functional aspects of the disease in a form suitable for downstream inference. The LLM is then prompted to identify drugs aligned with the extracted mechanisms. The prompt includes the disease context, the specified mechanism, and its role in disease pathogenesis, and requests a list of up to 50 single-agent drugs capable of modulating the mechanism. For each drug, the model provides a brief explanation describing how it affects the relevant pathway, target, or biological process. Combination therapies are explicitly excluded to maintain focus on single-agent interventions.

Together, these steps enable the LLM to generate candidates informed by disease-specific mechanistic reasoning. This approach expands the candidate pool beyond what is obtainable through KG embeddings alone, particularly for diseases with sparse graph connectivity or mechanisms not fully captured in structured resources.

*Knowledge Graph Embedding (KGE) Model: Hierarchy-Aware Knowledge Graph Embedding (HAKE)*

To model structured relationships among drugs, diseases, genes, and other biomedical entities, DrugKLM employs the Hierarchy-Aware Knowledge Graph Embedding (HAKE) model[19] trained directly on PrimeKG[20]. PrimeKG integrates curated biomedical knowledge from multiple authoritative sources and provides a relational structure suitable for embedding-based inference.

In HAKE, each entity and relation is represented using a modulus (radial) component encoding its hierarchical depth (i.e., from general to specific) and a phase (angular) component encoding semantic similarity.

For a triple $(h, r, t)$, HAKE computes a relevance score:

$$s(h,r,t) = -\|h_{mod} \circ r_{mod} - t_{mod}\|_2 - \lambda \cdot \left\| sin\left(\frac{h_{phase} + r_{phase} - t_{phase}}{2}\right) \right\|_1$$

where ∘ denotes element-wise multiplication and $\lambda$ balances the contribution of hierarchical structure and semantic similarity.

For drug repurposing task, the input is a disease entity $h$, and the objective is to identify drug entities $t$ likely connected through the indication relation $r$. Candidate drugs are ranked according to their HAKE scores, which reflect structural proximity within the knowledge graph. The KGE module returns the top 100 drug candidates with the highest relevance scores.

*Literature-Weighted Knowledge Graph Embedding (KGwE)*

To incorporate literature support into the knowledge graph embedding model, we extend HAKE with a literature-guided weighting mechanism. Each triple $(h, r, t)$ is assigned an initial weight $w_{(h,r,t)}^{(0)}$:

$$w_{(h,r,t)}^{(0)} = 1 + log\left(1 + N_{(h,r,t)}\right)$$

where $N_{(h,r,t)}$ denotes the number of PubMed articles supporting the triple. Article counts were obtained using the PubTator API (https://www.ncbi.nlm.nih.gov/research/pubtator3/api).

Rather than remaining static, each literature-guided weight $w_{(h,r,t)} \in [0,1]$ is modeled as a learnable scalar representing confidence in the triple, and is jointly optimized updated with the embedding parameters during training. This dynamic weighting allows the model to prioritize literature-supported relationships while retaining flexibility to adjust weights based on structural signals.

Training uses a weighted binary cross-entropy (BCE) loss over positive and negative samples. For each observed triple $(h, r, t)$, a corresponding negative sample $(h', r, t')$ is generated by randomly corrupting either the head or tail entity. Negative samples that already exist in the knowledge graph are filtered out to avoid false negatives.

The weighted loss $L_w$ is defined as:

$$L_w \ = \frac{1}{\sum w_{(h,r,t)}} \sum_{(h,r,t)} w_{(h,r,t)} \cdot \left[-log\ \sigma\left(s(h,r,t)\right) - log\ \sigma\left(-s(h',r,t')\right)\right]$$

where $s(h, r, t)$ is the HAKE scoring function and $\sigma(\cdot)$ denotes the sigmoid function, which maps scores to probabilities in $[0,1]$.

This KGwE model enables the embedding space to reflect both structural knowledge and degree of literature support, integrating relation inference with evidence strength from biomedical publications.

**Evidence Exploration**

DrugKLM assembles a multi-evidence profile for each drug–disease pair by integrating entity-level and pathway-level biomedical signals. Specifically, it aggregates drug-, gene-, and pathway-centered evidence derived from the knowledge graph, biomedical literature, and perturbational transcriptomic data.

*Drug-Level Evidence*

To collect drug-level evidence, DrugKLM integrates structured graph signals with text-derived contextual evidence.

For each drug–disease pair, the system retrieves shortest paths in PrimeKG that represent relational patterns such as shared targets, implicated pathways, or intermediary phenotypes. These paths encode potential mechanistic explanations linking the drug and disease.

To ensure biological interpretability, the *synergistic interaction* relation is excluded from the shortest-path search. Synergistic interactions describe effects arising only from combination therapy and therefore do not represent mechanisms attributable to a single agent. Including such edges could introduce combination–specific shortcuts that obscure mechanistic coherence in single-drug repurposing.

To prioritize paths, we define a path-based HAKE scoring function that evaluates semantic consistency across the sequence of entities and relations. Each edge corresponds to a biological relation in PrimeKG (e.g., indication, enzyme, associated with).

First, the HAKE score for each triple is normalized to a probabilistic scale [0,1]:

$$s'(h,r,t) = \frac{1}{1+\exp\left(-\frac{s(h,r,t)-\mu}{\sigma}\right)}$$

where $\mu = 350$ (the offset) and $\sigma = 100$ (the scaling factor).

For a multi-hop path $P = (e_0 \rightarrow e_1 \rightarrow \cdots \rightarrow e_n)$, the overall path score is computed using the geometric mean of the normalized edge scores:

$$p(e_0, e_n) = \left(\prod_{i=0}^{n-1} s'^{(e_i, r_i, e_{i+1})}\right)^{1/n-1}$$

The geometric mean ensures comparability across paths of different lengths and penalizes longer or noisier paths by requiring consistently high scores at each step. This favors shorter, semantically coherent paths and reflects increased uncertainty in extended reasoning chains.

For each drug-disease pair, up to ten highest-scoring shortest paths are retained to construct the knowledge subgraph.

DrucKLM retrieves literature evidence using the PubTator3 API. For each drug–disease pair, a co-occurrence query is submitted, and up to five highlighted PubMed snippets mentioning both entities are collected. These snippets provide concise contextual signals, including references to prior studies, clinical findings, or indirect associations reported in biomedical literature.

DrugKLM incorporates regulatory context by extracting relevant sections from FDA-approved drug labeling. For each candidate drug, the system retrieves statements from sections including indications and usage, mechanism of action, contraindications, and warnings and precautions. These provide authoritative clinical and regulatory information relevant to repurposing assessment.

Together, knowledge graph paths, knowledge graph paths, literature snippets and FDA labeling form the drug-level evidence used in downstream LLM-based reasoning. The KG subgraphs capture curated mechanistic relationships, literature snippets provide contextual natural-language signals, and FDA labels contribute regulatory and clinical grounding. This integrated evidence supports biologically coherent and clinically contextualized drug repurposing predictions.

*Gene-Level Evidence*

DrugKLM collects gene-level evidence to characterize genes appearing in KG subgraphs or literature-derived signals and to assess their mechanistic contributions to the drug–disease relationship. This module integrates structured KG associations, literature-curated annotations, and embedding-based proximity signals to quantify a gene's relevance to both the drug and the disease.

For each drug–gene pair, DrugKLM aggregates evidence from three complementary categories:

1. Literature snippets: Drug–gene co-mention sentences are retrieved using the PubTator3 API and cleaned to produce concise snippets reflecting co-occurrence in biomedical literature.

2. External resource evidence: Multiple curated and data-driven resources capture orthogonal signals of drug–gene association, including:
   a. CTD chemical-gene interactions, providing curated interaction types with supporting PMIDs.
   b. PubTator3 machine-extracted relations, supplying automatically identified relation types and associated publication counts.
   c. DGIdb[40] curated interactions, reporting pharmacologic interaction types (e.g., inhibitor, activator) and regulatory attributes such as FDA approval or anti-neoplastic annotation.
   d. SigCom-LINCS perturbational signatures, capturing the direction and magnitude of drug-induced transcriptomic changes for the gene.
3. Knowledge graph paths: DrugKLM retrieves mechanistic paths from PrimeKG connecting Disease → Gene and Drug → Gene. Up to the ten highest-scoring shortest paths are retained, outlining plausible mechanistic routes linking the gene to both disease biology and drug action.

After assembling the gene-level evidence, DrugKLM employs an LLM to prioritize and summarize the most mechanistically relevant genes for each drug–disease pair. Up to ten genes are selected, and the model generates concise summaries integrating heterogeneous evidence into a unified narrative. These summaries describe how each gene contribute to disease biology and may mediate or modulate the drug's therapeutic mechanism.

*Pathway-Level Evidence*

To characterize pathway-level mechanisms linking each drug to the disease context, DrugKLM integrates perturbational signatures from SigCom-LINCS L1000 chemical perturbation data with variant-aware cell line annotations and drug sensitivity measurements.

For a given disease description, representative cell lines are identified from Cellosaurus, optionally restricted to those harboring specified driver variants (e.g., *BRAF* V600E for "BRAF V600E–mutated cutaneous melanoma"). For each candidate drug profiled in these cell lines, LINCS-derived up- and down-regulated genes are combined with dose information and IC50 values from GDSC[41] to construct disease-relevant drug signatures. Genes are ranked reflecting both transcriptional effect size and drug potency (see next section).

The top 200 up-regulated and top 200 down-regulated genes are submitted to Enrichr[42] for gene set enrichment analysis. The resulting enriched pathways and functional annotations, together with their directionality (up- or down-regulated), constitute the pathway-level evidence used to assess mechanistic coherence for each drug–disease pair.

*Construction of Drug-Specific Gene Signatures*

For each drug, gene-level perturbation evidence was obtained from the LINCS L1000 chemical perturbation dataset. Each LINCS signature includes dose, direction of regulation (up or down), and drug sensitivity metrics (IC50). To construct a per-drug transcriptional signature, we compute a weighted score for each gene by combining dose-dependent expression changes with IC50-based potency.

For each LINCS signature, we compute an IC50 weight and a dose weight. The IC50 weight is defined as:

$$w_{\mathrm{IC50}} = min\left(\frac{1}{IC50}, 1.0\right)$$

so that more potent drugs (lower IC50) receive higher weight, with a cap at 1.0. Dose values are converted to micromolar units and mapped to dose weight:

$$w_{\text{Dose}} = e^{-k\cdot\ln(1+\text{Dose})}$$

with $k = 0.5$ by default.

For each gene $g$, we aggregate evidence across all LINCS signatures for that drug. Let $U_g$ and $D_g$ denote the sets of signatures in which $g$ is up- or down-regulated, respectively. We compute:

$$\text{IC50Score}_g = \frac{\sum_{i\in U_g} w_{IC50,i} - \sum_{j\in D_g} w_{IC50,j}}{|U_g| + |D_g|}, \text{DoseScore}_g = \frac{\sum_{i\in U_g} w_{Dose,i} - \sum_{j\in D_g} w_{Dose,j}}{|U_g| + |D_g|}$$

Both scores are normalized to $[-1,1]$, and a composite score is computed:

$$\text{FinalScore}_g = \alpha \cdot \text{DoseScore}_{g,\text{norm}} + (1-\alpha) \cdot \text{IC50Score}_{g,\text{norm}}$$

with default weight $\alpha = 0.2$. Genes with $\text{FinalScore}_g > 0$ are classified as up-regulated and those with $\text{FinalScore}_g > 0$ as down-regulated. The top 200 genes in each direction, ranked by $|FinalScore_g|$, are retained as the drug-specific up- and down-regulated gene sets used in downstream analyses.

**Chain-of-Thought Reasoning**

Given drug-, gene-, and pathway-level evidence, DrugKLM incorporates a chain-of-thought (CoT) reasoning procedure to improve interpretability and mitigate hallucination risks.

The LLM is instructed to evaluate each candidate drug–disease pair using evidence derived from the knowledge graph and biomedical literature. The CoT procedure consists of three steps:

1. *Supporting evidence summary:* Summarizes retrieved evidence (literature snippets, FDA labeling excerpts where available, and knowledge graph paths) to assess therapeutic relevance.
2. *Risk and limitation analysis:* Identifies potential concerns – such as toxicity, off-target effects, limited specificity, or conflicting evidence – based on the collected evidence.
3. *Confidence scoring:* Assigns a confidence score in $[0, 1]$ based on the integrated reasoning context. This score reflects the overall strength of the repurposing hypothesis and supports expert triage and prioritization.

**Survival-Based Estimation of Drug–Disease Hazard Ratios**

To evaluate whether drug-associated transcriptional signatures are linked to patient survival, we estimated hazard ratios (HRs) for drug–disease pairs using TCGA gene expression and clinical survival data. This analysis does not assume that TCGA patients received the evaluated drugs. Instead, it tests whether the biological activity represented by a drug's LINCS-derived perturbational signature is directionally associated with overall survival within each cancer type.

*Computation of Enrichment Scores in TCGA Samples*

To quantify the similarity between each TCGA tumor and each drug's perturbational signature, we applied single-sample gene set enrichment analysis (ssGSEA)[43] to the TCGA expression matrix. For each drug and patient sample, ssGSEA generated enrichment score (ES) and its normalized enrichment scores (NES) for both the up-regulated and down-regulated gene sets.

*Stratification of TCGA Patients*

For each drug-disease pair, patients were stratified into High and Low groups based on the distribution of NES values. A tertile-based approach was used: only the top and bottom one-third of samples were retained as the High and Low groups, respectively.

To ensure stable survival estimation, analyses were restricted to drug–disease pairs in which both groups contained at least 10 samples and then corresponding cancer type had at least 3 observed survival events.

*Survival Association and Validation*

For each eligible drug–disease, a univariable Cox proportional hazards model was fitted using a binary indicator (High vs. Low NES). The resulting hazard ratio (HR) was used as the survival-associated estimate for downstream analyses. An HR < 1 indicates that tumors more strongly resembling the drug's signature are associated with improved survival, whereas HR > 1 indicates the opposite trend.

To evaluate clinical alignment, we computed the Spearman correlation between HR values and DrugKLM confidence scores across evaluated TCGA cancer types. This analysis tests whether higher model-assigned confidence is associated with more favorable survival-linked transcriptional patterns.

**ClinicalTrials.gov and PubMed–Based Drug Categorization**

To determine the clinical evidence level of each predicted drug–disease pair, DrugKLM integrates data from ClinicalTrials.gov and PubMed.

For each pair, drug and disease terms are matched against locally indexed ClinicalTrials.gov intervention and condition records to identify relevant trial identifiers (NCT IDs). For each matched trial, metadata—including trial phase, recruitment status, availability of posted results, and last update date—are retrieved via the ClinicalTrials.gov API.

In parallel, PubMed is queried to identify publications associated with matched NCT identifiers, as well as drug- and disease- specific literature when no trials are found.

All retrieved trial- and literature-level evidence is aggregated and provided to a GPT-based classifier, which assigns each drug–disease pair to one of 17 predefined clinical evidence stages (in Supplementary Table S5), ranging from FDA-approved indications and phase-specific trial outcomes to preclinical or insufficient evidence.

**Blinded Expert Annotation of Top-Ranked Candidates**

To assess whether model rankings corresponded to clinical evidence stage, we conducted blinded human curation of the top 30 drugs ranked by DrugKLM and TxGNN across five cancers (liver cancer, colon cancer, pancreatic cancer, melanoma, and glioblastoma). Curation was performed independently by clinicians and cancer researchers with formal biomedical training and experience in evaluating clinical and translational evidence, with at least one curator assigned per disease. Prior to annotation, curators participated in a brief guideline-based discussion to standardize interpretation of predefined evidence categories. Each drug–disease pair was classified according to the clinical evidence stage defined in Supplementary Table S7. Classification was based on reported clinical and preclinical evidence derived from ClinicalTrials.gov records, peer-reviewed biomedical literature, and FDA approval status. Curated data are provided in Supplementary Table S2.

## Code availability

The code is freely available: https://github.com/ncbi-nlp/DrugKLM.

## Human Ethics and Consent to Participate declarations

not applicable

## Acknowledgemensts

This research was supported by the Division of Intramural Research (DIR) of the National Library of Medicine (NLM), National Institutes of Health. The authors thank the generous contributions from the Florence D. and Irving J. Sherman MD Charitable Foundation Trust, the Reza Khatib MD Brain Tumor Center, the Venable Foundation, and the Leonard Attman family for funding this work.

## References

1. Matsumoto, N., *et al.* ESCARGOT: an AI agent leveraging large language models, dynamic graph of thoughts, and biomedical knowledge graphs for enhanced reasoning. *Bioinformatics* **41**, btaf031 (2025).
2. Abdullahi, T., *et al.* K-paths: Reasoning over graph paths for drug repurposing and drug interaction prediction. in *Proceedings of the 31st ACM SIGKDD Conference on Knowledge Discovery and Data Mining V. 2* 5-16 (2025).
3. Huang, K., *et al.* A foundation model for clinician-centered drug repurposing. *Nature Medicine* **30**, 3601-3613 (2024).
4. Wu, J., *et al.* DrugSim2DR: systematic prediction of drug functional similarities in the context of specific disease for drug repurposing. *GigaScience* **12**, giad104 (2023).
5. Shao, M., Jiang, L., Meng, Z. & Xu, J. Computational drug repurposing based on a recommendation system and drug–drug functional pathway similarity. *Molecules* **27**, 1404 (2022).
6. He, H., *et al.* Computational drug repurposing by exploiting large-scale gene expression data: Strategy, methods and applications. *Computers in biology and medicine* **155**, 106671 (2023).
7. Kwee, I., Martinelli, A., Khayal, L.A. & Akhmedov, M. metaLINCS: an R package for meta-level analysis of LINCS L1000 drug signatures using stratified connectivity mapping. *Bioinformatics Advances* **2**, vbac064 (2022).
8. Nunes, S., Badreddine, S. & Pesquita, C. Rewarding explainability in drug repurposing with knowledge graphs. *arXiv preprint arXiv:2509.02276* (2025).
9. Labrak, Y., *et al.* Biomistral: A collection of open-source pretrained large language models for medical domains. *arXiv preprint arXiv:2402.10373* (2024).

10. Wang, Z.P., *et al.* Drug repurposing for Alzheimer's disease using a graph-of-thoughts based large language model to infer drug-disease relationships in a comprehensive knowledge graph. *BioData Mining* **18**, 51 (2025).
11. Huang, L.-C., *et al.* DrugReX: an explainable drug repurposing system powered by large language models and literature-based knowledge graph. *Research Square*, rs. 3. rs-6728958 (2025).
12. Safaei, A.A., Saboori, P., Ramezani, R. & Nematbakhsh, M. KGLM-QA: A Novel Approach for Knowledge Graph-Enhanced Large Language Models for Question Answering. in *2024 15th International Conference on Information and Knowledge Technology (IKT)* 234-240 (IEEE, 2024).
13. Liu, S., *et al.* Drugagent: Automating ai-aided drug discovery programming through llm multi-agent collaboration. *arXiv preprint arXiv:2411.15692* (2024).
14. Zhang, F., Zhao, Y., Zhang, W. & Lai, L. BioScientist Agent: Designing LLM-Biomedical Agents with KG-Augmented RL Reasoning Modules for Drug Repurposing and Mechanistic of Action Elucidation. *bioRxiv*, 2025.2008. 2008.669291 (2025).
15. Wei, J., *et al.* DrugReAlign: a multisource prompt framework for drug repurposing based on large language models. *BMC biology* **22**, 226 (2024).
16. Gao, S., *et al.* TxAgent: An AI agent for therapeutic reasoning across a universe of tools. *arXiv preprint arXiv:2503.10970* (2025).
17. Linehan, W.M. & Ricketts, C.J. The Cancer Genome Atlas of renal cell carcinoma: findings and clinical implications. *Nature Reviews Urology* **16**, 539-552 (2019).
18. Zarin, D.A., Tse, T., Williams, R.J., Califf, R.M. & Ide, N.C. The ClinicalTrials. gov results database—update and key issues. *New England Journal of Medicine* **364**, 852-860 (2011).
19. Zhang, Z., Cai, J., Zhang, Y. & Wang, J. Learning hierarchy-aware knowledge graph embeddings for link prediction. in *Proceedings of the AAAI conference on artificial intelligence*, Vol. 34 3065-3072 (2020).
20. Chandak, P., Huang, K. & Zitnik, M. Building a knowledge graph to enable precision medicine. *Scientific Data* **10**, 67 (2023).
21. Davis, A.P., *et al.* Comparative toxicogenomics database's 20th anniversary: update 2025. *Nucleic acids research* **53**, D1328-D1334 (2025).
22. Wei, C.-H., *et al.* PubTator 3.0: an AI-powered literature resource for unlocking biomedical knowledge. *Nucleic Acids Research* **52**, W540-W546 (2024).
23. Evangelista, J.E., *et al.* SigCom LINCS: data and metadata search engine for a million gene expression signatures. *Nucleic acids research* **50**, W697-W709 (2022).
24. Iorio, F., *et al.* A landscape of pharmacogenomic interactions in cancer. *Cell* **166**, 740-754 (2016).
25. Zhou, H., *et al.* MEDICASCY: a machine learning approach for predicting small-molecule drug side effects, indications, efficacy, and modes of action. *Molecular pharmaceutics* **17**, 1558-1574 (2020).
26. Himmelstein, D.S., *et al.* Systematic integration of biomedical knowledge prioritizes drugs for repurposing. *elife* **6**, e26726 (2017).
27. Rapisuwon, S., *et al.* Systemic therapy for mucosal, acral, and uveal melanoma. in *Cutaneous Melanoma* 1301-1335 (Springer, 2020).
28. Grasso, C.S., *et al.* Conserved interferon-γ signaling drives clinical response to immune checkpoint blockade therapy in melanoma. *Cancer cell* **38**, 500-515. e503 (2020).
29. Griss, J., *et al.* B cells sustain inflammation and predict response to immune checkpoint blockade in human melanoma. *Nature communications* **10**, 4186 (2019).
30. Birth, D., Kao, W.-C. & Hunte, C. Structural analysis of atovaquone-inhibited cytochrome bc 1 complex reveals the molecular basis of antimalarial drug action. *Nature communications* **5**, 4029 (2014).

31. Lee, J.S., *et al.* Synthetic lethality-mediated precision oncology via the tumor transcriptome. *Cell* **184**, 2487-2502. e2413 (2021).
32. Riaz, N., *et al.* Tumor and microenvironment evolution during immunotherapy with nivolumab. *Cell* **171**, 934-949. e916 (2017).
33. Hugo, W., *et al.* Genomic and transcriptomic features of response to anti-PD-1 therapy in metastatic melanoma. *Cell* **165**, 35-44 (2016).
34. Rooney, M.S., Shukla, S.A., Wu, C.J., Getz, G. & Hacohen, N. Molecular and genetic properties of tumors associated with local immune cytolytic activity. *Cell* **160**, 48-61 (2015).
35. Abdullahi, T., *et al.* K-paths: Reasoning over graph paths for drug repurposing and drug interaction prediction. *arXiv preprint arXiv:2502.13344* (2025).
36. Dong, X., *et al.* Personalized prediction of anticancer potential of non-oncology drugs through learning from genome derived molecular pathways. *NPJ Precision Oncology* **9**, 36 (2025).
37. Song, H., Bang, D., Koo, B., Kim, S. & Lee, S. LLM-Integrated Representative Path Selection for Context-Aware Drug Repurposing on Biomedical Knowledge Graphs. in *NeurIPS 2025 2nd Workshop on Multi-modal Foundation Models and Large Language Models for Life Sciences*.
38. More, V., *et al.* TheraMind: A Multi-LLM Agent for Accelerating Drug Repurposing in Lung Cancer via Case Report Mining. (2025).
39. Jin, Q., *et al.* Medcpt: Contrastive pre-trained transformers with large-scale pubmed search logs for zero-shot biomedical information retrieval. *Bioinformatics* **39**, btad651 (2023).
40. Cannon, M., *et al.* DGIdb 5.0: rebuilding the drug–gene interaction database for precision medicine and drug discovery platforms. *Nucleic acids research* **52**, D1227-D1235 (2024).
41. Yang, W., *et al.* Genomics of Drug Sensitivity in Cancer (GDSC): a resource for therapeutic biomarker discovery in cancer cells. *Nucleic acids research* **41**, D955-D961 (2012).
42. Kuleshov, M.V., *et al.* Enrichr: a comprehensive gene set enrichment analysis web server 2016 update. *Nucleic acids research* **44**, W90-W97 (2016).
43. Barbie, D.A., *et al.* Systematic RNA interference reveals that oncogenic KRAS-driven cancers require TBK1. *Nature* **462**, 108-112 (2009).

# Supplementary materials:

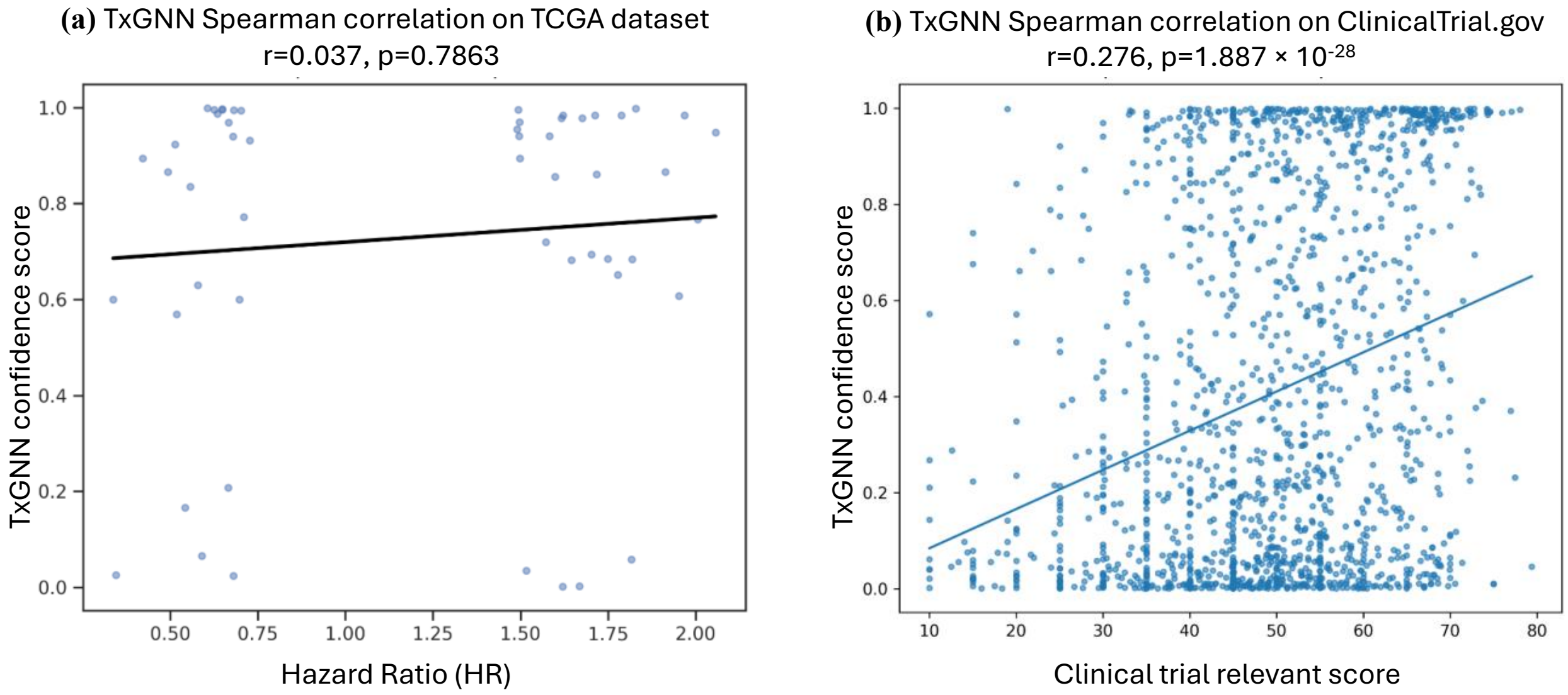

Fig. S1 Relevance scoring of TxGNN using TCGA survival outcomes and ClinicalTrials.gov trial statuses. (a) Correlation between TxGNN confidence scores and TCGA dataset survival relevance across drug–disease pairs. Each point represents one disease-drug pair, with the fitted line indicating the overall trend. (b) Correlation between TxGNN confidence scores and ClinicalTrials.gov trial relevance score.

**Clinical Validation of DrugKLM Scores: Trial-Aligned Signal**

In addition to the clinical validation on TCGA patient survival data, we further evaluated whether DrugKLM confidence scores align with clinical evidence by analyzing drug–disease pairs recorded in ClinicalTrials.gov. An LLM-based AI reviewer, guided by a scoring prompt (Fig. S4) that was reviewed by a cancer biologist, assessed mechanistic rationale, preclinical evidence, early clinical signals, class-level validation, PK/PD suitability, safety and tolerability, trial design quality, and the importance of the target drug within the study. Fig. S1a illustrates the relationship between the DrugKLM confidence score and the clinical relevance score across all evaluated drug-disease pairs with at least one associated clinical trial study. Each dot represents an individual pair, and the positive correlation demonstrates that higher DrugKLM scores correspond to higher clinical relevance. DrugKLM's confidence scores showed a strong and statistically significant overall correlation with the independently generated clinical relevance score ($r = 0.512$, $p = 5.099 \times 10^{-218}$, shown in Fig. S2a). Under the same analysis, TxGNN also exhibited a statistically significant but substantially weaker association ($r = 0.276$, $p = 1.887 \times 10^{-28}$, shown in Fig. S1b). Although this trial-text–based evaluation serves as an auxiliary validation, the findings are consistent with the TCGA survival analysis and further support the clinical relevance of DrugKLM. The confidence score is intended as an ordinal prioritization signal. It is generated under a fixed scoring rubric, which naturally produces semi-discrete values and is appropriate for rank-based evaluation. Fig. S2b presents the corresponding comparison using clinical relevance scores derived from AI-reviewed ClinicalTrials.gov studies. Here, pairs with greater clinical relevance again show stronger alignment with the DrugKLM axis. Taken together, these distributions show that DrugKLM more closely tracks both observational and interventional clinical evidence, indicating a stronger capacity to prioritize drug candidates with therapeutic potential. As shown in Fig. S2c, removal of any single evidence source resulted in a consistent reduction in alignment with ClinicalTrials.gov clinical-trial evidence. This trend is consistent with the TCGA survival outcomes.

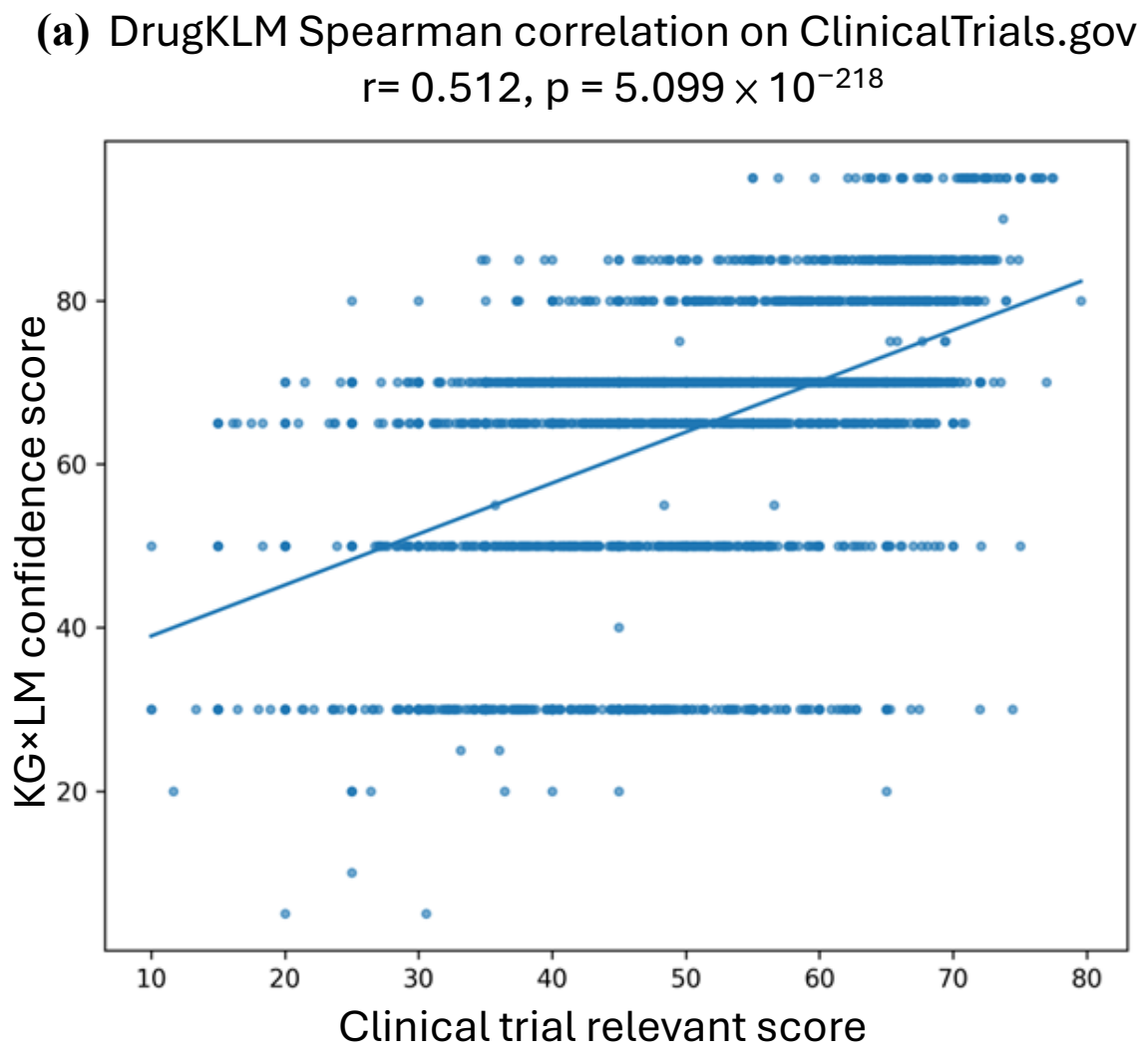

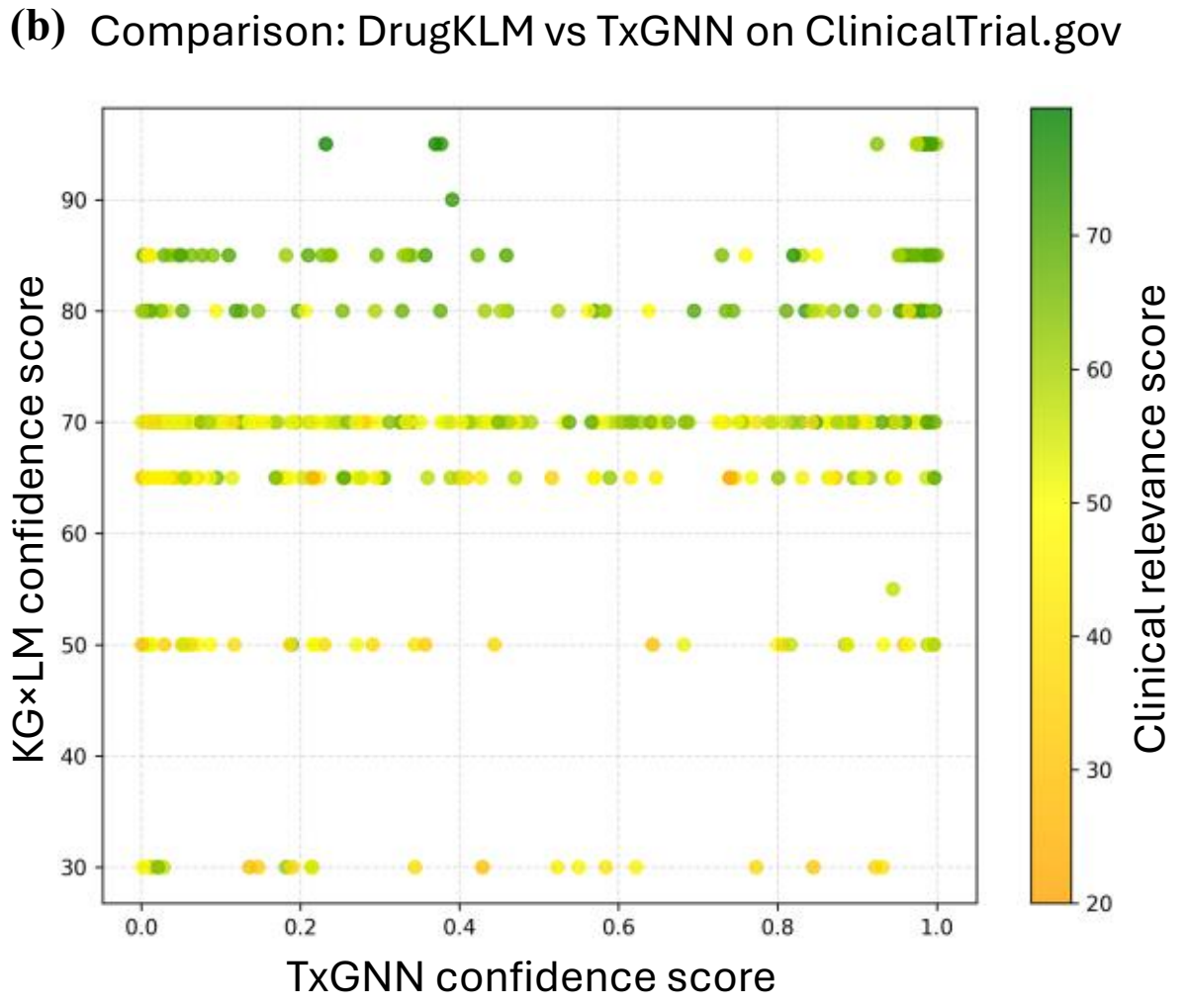

**(c)** Leave-one-evidence-out ablation analysis on ClinicalTrial.gov

| | Spearman r | p-value |
|---|---|---|
| DrugKLM | 0.512 | 5.10E-218 |
| DrugKLM w/o Drug Evidence | 0.498 | 5.15E-201 |
| DrugKLM w/o Gene Evidence | 0.500 | 2.64E-197 |
| DrugKLM w/o Pathway Evidence | 0.506 | 9.91E-203 |
| DrugKLM w/o CoT (using heuristic rule for scoring) | 0.319 | 6.68E-188 |

Fig. S2. Clinical validation of DrugKLM using ClinicalTrials.gov evidence. (a) Relevance between DrugKLM confidence scores and clinical relevance scores generated by an automated AI reviewer analyzing ClinicalTrials.gov trials. Each point represents a drug–disease pair with at least one associated clinical trial. Higher DrugKLM scores correspond to higher clinical relevance. (b) Direct comparison of DrugKLM and TxGNN confidence scores for drug–disease pairs. Dots are colored by clinical relevance. Pairs with higher clinical relevance cluster more strongly along the DrugKLM axis, indicating closer alignment with interventional clinical evidence. (c) Leave-one-evidence-out ablation analysis evaluating alignment with ClinicalTrials.gov evidence. Removal of any single evidence source reduces the Spearman correlation.

```
You are evaluating whether [Drug] is a promising repurposing candidate for [Disease]
based on the evidence provided in the Input section.

This is a FIXED HEURISTIC RULE-BASED BASELINE.
Do NOT perform reasoning, interpretation, or explanation.
```

Input Evidence:
1) Disease–Drug Evidence
2) Aggregated Disease–Gene and Drug–Gene Evidence
3) GSEA pathway evidence
4) Case JSON for disease context

[subtype_statements]

Scoring Rules (apply independently):
Disease–Drug evidence:
- If direct disease–drug evidence is supported by clinical trial reports
  or FDA-approved indications → add 40 points.
- If disease–drug evidence is indirect or preclinical only
  (e.g., cell line or animal studies) → add 20 points.

Gene-level evidence:
- If multiple disease-relevant genes with known mechanistic links
  are supported by literature or KG evidence → add 30 points.
- If limited or indirect gene-level associations are present → add 15 points.

GSEA pathway evidence:
- If statistically significant disease-relevant pathways
  (e.g., FDR < 0.05) are identified → add 20 points.
- If pathways are nominally significant or generic → add 10 points.

FDA approval:
- If FDA approval exists for any indication → add 10 points.

Constraints:
- Apply rules exactly as written.s
- Do NOT modify weights.
- Do NOT reconcile conflicting evidence.
- Do NOT perform mechanistic or clinical reasoning.

After scoring:
- Cap overall_confidence_score at 100.
- Map the score to confidence levels using predefined ranges.

Output Constraints:
- Populate the required JSON fields only.
- rationale_bullets MUST list ONLY rule triggers (no explanation).

Output:
Return ONLY the following JSON structure:

```
{
  "Disease": "[Disease]",
  "Drug": "[Drug]",
  "verdict": {
    "overall_confidence_level": "Very High|High|Moderately High|Moderate|Moderately Low|Low|Very Low",
    "subtype_confidence_level": "Very High|High|Moderately High|Moderate|Moderately Low|Low|Very Low|N/A",
    "subtype": "[subtypes]||N/A",
    "risk_level": "Very High|High|Moderately High|Moderate|Moderately Low|Low|Very Low",
    "overall_confidence_score": [0-100],
    "subtype_confidence_score": [0-100],
    "risk_score": [0-100],
    "FDA_status": "FDA-approved for [Disease]|FDA-approved for other indications but not for [Disease]|N/A"
  },
  "rationale_bullets": [
    "Rule triggered: Disease–Drug evidence (+40).",
    "Rule triggered: Gene-level evidence (+30)."
  ]
}
```

Input:
Case JSON:
[JSON_Input]

Evidence Payload:
[input]

Fig. S3. Prompt design for a non-CoT baseline that applies a structured, domain-informed evaluation.

You are an expert clinical reviewer specializing in translational oncology, immunotherapy, and early-phase drug development.
Your task is to analyze the clinical trial study provided at the end of this prompt (after the keyword INPUT:) and produce a structured JSON output.
The goal is to evaluate the scientific confidence of the investigational drug: "[Drug]" against the disease: "[Disease]" based on defined criteria.

The analysis must follow these rules:

1. Output must be valid JSON.

2. Each score must be between 0 and 100.
3. All reasoning must be grounded in:

   * The provided study information
   * Established scientific knowledge of oncology drug development
4. Do not use bullets or symbols that are not on a standard keyboard.
5. Include both:
   a) a detailed explanation for each scoring category
   b) the numeric score
6. Use the evaluation features and importance levels listed below.

---

EVALUATION FEATURES AND THEIR IMPORTANCE

Mechanistic rationale (Importance: Highest)

* How strongly the drug mechanism links to the disease biology.
* Whether the target is known to be relevant.
* Whether similar mechanisms have proven successful.

Preclinical evidence (Importance: High)

* Strength of in vitro and in vivo activity.
* Predictiveness of models used.
* Dose response and translational validity.

Early clinical signals (Importance: Highest)

* Evidence directly from this study or similar patients, including objective responses, duration, pharmacodynamic activity, or any other measurable antitumor effect.
* If the trial is Phase I or Phase II, has enrolled patients, more than two years have passed since primary completion, and no study results, publications, abstracts, or press releases are available, the clinical evidence should be interpreted as weak.
* If a study remains Not Yet Recruiting without enrollment, it provides no clinical evidence, and confidence is low because no human data are available.

Class validation (Importance: High)

* Whether other drugs with the same target/pathway are validated.
* External evidence from approved or late-stage agents.

PK/PD suitability (Importance: Medium)

* Exposure, half-life, receptor occupancy, drug distribution.
* Whether [Drug] reaches and maintains levels needed for activity.

Safety and tolerability (Importance: Medium)

* Expected toxicities based on mechanism.
* DLTs, immune-related AEs, combinational toxicity patterns.

Trial design quality (Importance: Low-medium)

* Whether the study design allows meaningful signal detection.
* Strength of endpoints and inclusion criteria.

Manufacturing readiness (Importance: Low)

* Whether [Drug] has known stability, scalability, CMC feasibility.

Competitive landscape (Importance: Lowest)

* Availability of alternative therapies.
* Whether unmet need influences value but not biological confidence.

Target drug importance (Importance: High)

* Indicates how central the [Drug] is to the study outcome.
* Higher scores mean the [Drug] is the primary experimental agent and the results are directly informative for evaluating it.
* Lower scores apply when the [Drug] is only a combination partner, background therapy, or part of a control arm, making its individual contribution unclear.

Result status scoring rules (Importance: High)

* bad: 0 to 19. Study has available negative or non-efficacious results.
* completed_no_result: 20 to 34. The study is marked Completed for more than 12 months with no posted results.
* ongoing_long_term_incomplete: 35 to 49. If the study has no posted results AND exceeds normal development timelines, it MUST be classified as long_term_incomplete (35–49) even if it is still listed as Recruiting or Active.
    "this_year-start_year":2+ for Phase I/II
    "this_year-start_year":5+ for Phase III
* ongoing_in_reasonable_term: 50 to 79. Only if none of the above apply → ongoing_in_reasonable_term (50 – 79).
    "this_year-start_year":<2 for Phase I/II
    "this_year-start_year":<5 for Phase III
* good: 80 to 100. Study has available positive or clinically meaningful results.

---

SCORING GUIDELINES (0-100)

* 0 to 20: Very low confidence
* 21 to 40: Low confidence

* 41 to 60: Moderate confidence
* 61 to 80: High confidence
* 81 to 100: Very high confidence

---

Example output format (structure only):

```
{
	"study_summary": {
		"target_disease": "[Disease]", <- the variable ALWAYS refers to the user-provided Target_Disease
		"target_drug": "[Drug]", <- the variable ALWAYS refers to the user-provided Target_Drug
		"title": "...",
		"NCTID":"...",
		"drug_for_treatment": ["..", ".."],
		"indication": ""...,
		"result_summary_sentence": "...",
		"phase": "...",
		"risk_level": "...",
		"positive_or_negative": "",
		"population_summary": "",
		"has_results": true|false,
		"sample_size": [0-9]+,
		"start_year": [0-9]+,
		"this_year-start_year": [0-9]+,
		"complete year": [0-9]+
	},
	"scoring": {
		"mechanism": {
			"score": 0-100,
			"explanation": ""
		},
		"clinical_evidence": {
			"score": 0-100,
			"explanation": ""
		},
		"safety_predictability": {
			"score": 0-100,
			"explanation": ""
		},
		"trial_quality": {
			"score": 0-100,
			"explanation": ""
		},
		"result_status": {
			"score": 0-100,
```

```
				"category": "",	// "good" / "bad" / "completed_no_result" / 
"ongoing_long_term_incomplete" / "ongoing_in_reasonable_term"
				"explanation": ""
			}
			"target_drug_importance": {
				"score": 0-100,
				"role": "",				// "major" / "combination_partner" / 
"background_therapy" / "control"
				"explanation": ""
			},
			"overall_confidence": {
				"score": 0-100,
				"explanation": ""
			}
		}
}
```

---

TASK

1. Read the clinical trial study from INPUT:[Input]
2. Produce the JSON described above.
3. The JSON must be comprehensive and self-contained.
4. Avoid any special characters not available on a standard keyboard.
5. When assigning overall_confidence, the model must treat the score of result_status as a major determinant.

---

INPUT:
[Input]

Target_Disease:
[Disease]

Target_Drug:
[Drug]

Fig. S4. Scoring prompt for automated ClinicalTrials.gov–based relevance evaluation.